\documentclass{article}

\PassOptionsToPackage{numbers, compress}{natbib}

\usepackage[final]{neurips_2022}

\usepackage[utf8]{inputenc} 
\usepackage[T1]{fontenc}    
\usepackage{hyperref}       
\usepackage{url}            
\usepackage{amsfonts}       
\usepackage{nicefrac}       
\usepackage{microtype}      
\usepackage{times}
\usepackage{epsfig}
\usepackage{graphicx}
\usepackage{wrapfig}
\usepackage{amsmath}
\usepackage{amssymb}
\usepackage{algorithm,algorithmic}
\usepackage{multirow}
\usepackage{graphicx}
\usepackage[table,dvipsnames]{xcolor}
\usepackage{pifont}
\newcommand{\xmark}{\ding{55}}
\usepackage{booktabs}
\usepackage{colortbl}
\usepackage{comment}

\title{An Intelligent Modular Real-Time Vision-Based System for Environment Perception}

\vspace{-2em}
\author{%
	Amirhossein Kazerouni\thanks{Equal contribution} \\
	Iran University of Science and Technology \\
	\texttt{amirhossein477@gmail.com} \\ \And
 	Amirhossein Heydarian\footnotemark[1] \\
	Amirkabir University of Technology \\
	\texttt{ah.heydarian@aut.ac.ir} \\ \And
 	Milad Soltany\footnotemark[1] \\
	Iran University of Science and Technology \\
	\texttt{soltany.m.99@gmail.com} \\ \And
 	Aida Mohammadshahi\footnotemark[1]  \\
	University of Calgary \\
	\texttt{aida.mohammadshahi@ucalgary.ca} \\ \And
 	Abbas Omidi\footnotemark[1] \\
	University of Calgary \\
	\texttt{abbas.omidi@ucalgary.ca} \\ \And
 	Saeed Ebadollahi \\
	Iran University of Science and Technology \\
	\texttt{s\_ebadollahi@iust.ac.ir} \\
}

\begin{document}

\maketitle

\vspace{-1.5em}
\begin{abstract}
\vspace{-1em}
    A significant portion of driving hazards is caused by human error and disregard for local driving regulations; Consequently, an intelligent assistance system can be beneficial. This paper proposes a novel vision-based modular package to ensure drivers' safety by perceiving the environment. Each module is designed based on accuracy and inference time to deliver real-time performance. As a result, the proposed system can be implemented on a wide range of vehicles with minimum hardware requirements.
    Our modular package comprises four main sections: lane detection, object detection, segmentation, and monocular depth estimation. Each section is accompanied by novel techniques to improve the accuracy of others along with the entire system. 
    Furthermore, a GUI is developed to display perceived information to the driver. In addition to using public datasets, like BDD100K, we have also collected and annotated a local dataset that we utilize to fine-tune and evaluate our system. We show that the accuracy of our system is above 80\% in all the sections. Our code and data are available on \href{https://github.com/Pandas-Team/Autonomous-Vehicle-Environment-Perception}{\textcolor{NavyBlue} {GitHub}.}
\end{abstract}

\vspace{-1.5em}
\section{Introduction}
\vspace{-1em}
As the number of vehicles on the road has grown in recent years, traffic violations, accidents, and fatalities have increased considerably \cite{duivenvoorden2010relationship}. However, along with the growth in urban traffic, human error plays a significant role in the frequency of road casualties and offenses, in a way that careless driving and disregard for traffic signs account for more than 70\% of street accidents \cite{bener2005neglected}.
Hence, providing a realistic solution to improve driving accuracy and road safety could be highly beneficial. Artificial intelligence has advanced many fields, including the automotive industry \cite{DataInjection,solarbased,triobjective}. Some companies have offered autonomous vehicles as an alternative to human driving \cite{ingle2016tesla}, but they are still not commonly used by the public due to their high cost. 

Road lane detection is a significant part of the perception system to determine the vehicle's position and its desired trajectory on the road, which remains problematic in intelligent vehicles. The challenges of this problem include disordered, fading, and unreasonable road markings, along with various lighting conditions. Furthermore, road lines might be occluded by other cars and obstacles. Early lane detection methods include classical image processing techniques based on image edges, and the Hough transform \cite{kong2010general, liu2017road, panfilova2021fast, wang2021improved, sultana2021robust}. Despite their simplicity, these algorithms require environment-dependent hyperparameter adjustments. \cite{omidi2021embedded} presents a hybrid method based on object detection to improve lane detection accuracy. On the other hand, deep learning-based methods such as Point Instance Network (PINet) \cite{ko2021key} and LaneATT \cite{tabelini2021keep} offer precise and robust performance in multiple scenarios with different lighting conditions, and occlusions \cite{qu2021focus, liu2021condlanenet, abualsaud2021laneaf, haris2021lane}. As a result, deep-learning-based approaches are considered a better choice for practical applications.

Object Detection methods are used to detect various objects in the environment. Most detectors feed an image into an artificial neural network and output bounding box locations for each object present in the image and probability scores corresponding to each object class. The detector's head connecting to the backbone is usually one or two-stage. One-stage detectors, for instance, YOLO and SSD, are generally much faster than their two-stage counterparts \cite{jiao2019survey, bochkovskiy2020yolov4, redmon2016you}, while two-stage detectors, like the R-CNN family, tend to yield more outstanding accuracy scores \cite{liu2016ssd, lin2017focal, girshick2014rich, girshick2015fast, ren2015faster}. Our first use of object detection is detecting pedestrians and other vehicles, static and dynamic; this is done to avoid collisions with other objects on the road. Secondly, traffic indicators are detected to ensure vehicles follow the regulations on different roads \cite{de1997road}.

An intelligent system should accurately identify the road-sidewalk boundaries and measure its distance from detected objects to prevent accidents. In this case, pixel-wise classification of an image (semantic segmentation) offers impressive accuracy in determining the boundary by recognizing the sidewalk \cite{newell2016stacked, chen2018encoder, strudel2021segmenter, wu2021fully, localbinary2020deep, SVMclassification2017}. In addition, perspective transform approaches \cite{kim2019efficient, tuohy2010distance} that estimate the distance based on the warped image's pixel spacing and monocular depth estimation methods \cite{wang2020sdc, miangoleh2021boosting, chen2021attention, lee2022edgeconv} that measure the distance based on the depth map and camera parameters have shown promising results. However, the accuracy of perspective-transform-based approaches is highly dependent on the situation and cannot perform well in all conditions. Moreover, it is not a computationally cost-effective solution to measure the distance and detect the sidewalk individually through different models. For this purpose, models based on multi-task learning can be advantageous since they output both semantic segmentation and depth estimation in one model \cite{zhang2018joint, yang2018segstereo, chen2019towards}. These models simultaneously enhance the accuracy of segmentation and depth estimation through their mutual effect during training. For instance, SGDepth \cite{sgdepth} outputs the estimated depth map and semantic segmentation from a single input image.
\begin{figure}
	\centering
	\includegraphics[width=0.9\linewidth]{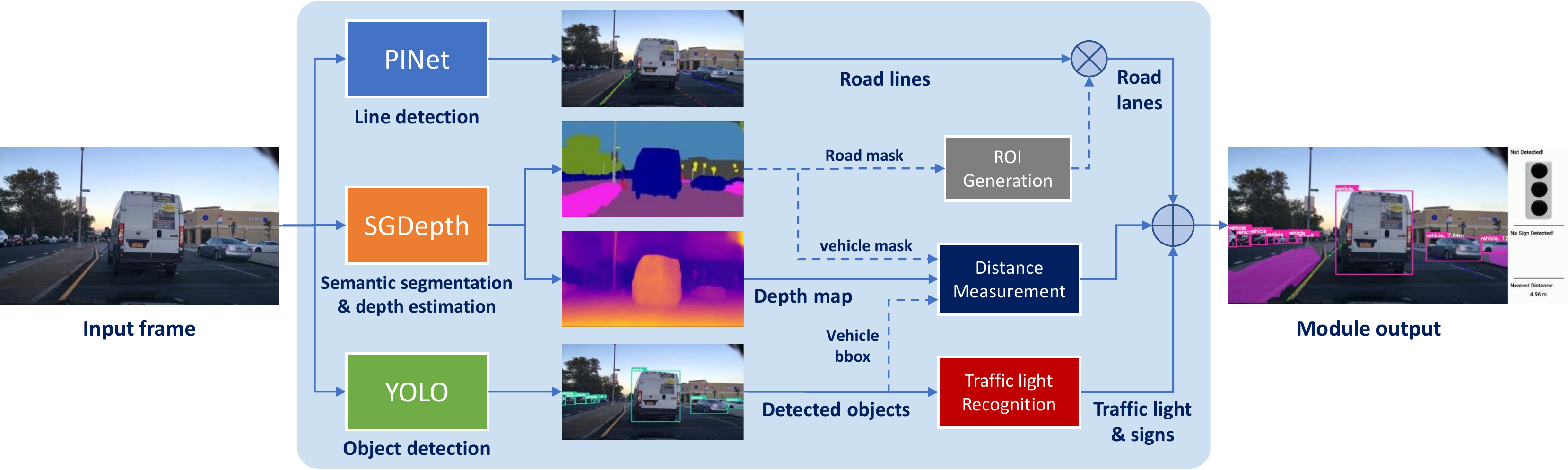}
	\vspace{-0.5em}
	\caption{\textbf{Overview of the proposed system.} Continuous lines represent system modules' main data path, and dashed lines denote the additional data, which aids in the accuracy of the system.}
	\label{fig:overall_diagramm}
	\vspace{-1.5em}
\end{figure}
\vspace{0.1em}
In this research, we have developed a comprehensive package for real-time environmental perception based on computer vision techniques to assist drivers in minimizing driving faults and violations.
To the best of our knowledge, we have provided a novel combination of traits, and only a few studies are available on the permutations of features mentioned in this paper. For instance, compared to the YOLOP\cite{wu2021yolop} network, this network lacks depth estimation and distance measurement, whereas our package is more extensive. There are two aspects to our real-time intelligent package: software and hardware. The software section consists of four main phases. For the road lane detection part, our module benefits from the PINet \cite{ko2021key} neural network, which is trained on the CULane \cite{shirke2019lane} dataset and fine-tuned on our collected local dataset. For the object detection section, we utilize YOLOv5 \cite{glenn_jocher_2022_6222936} for detecting vehicles, pedestrians, and traffic signs. In the third phase, we use the SGDepth \cite{sgdepth} network for segmentation tasks and recognizing the sidewalks. Finally, we introduce a novel approach for measuring the distance to the surrounding cars based on the monocular depth estimation output from the SGDepth. In terms of hardware, this module uses only one camera and a mid-range GPU, reducing costs and making it effortless to implement on various vehicles.

\vspace{-1.5em}
\section{Background}
\vspace{-1.0em}
\subsection{PINet}
\begin{figure}[!t]
\minipage{0.5\textwidth}
  \includegraphics[width=\linewidth]{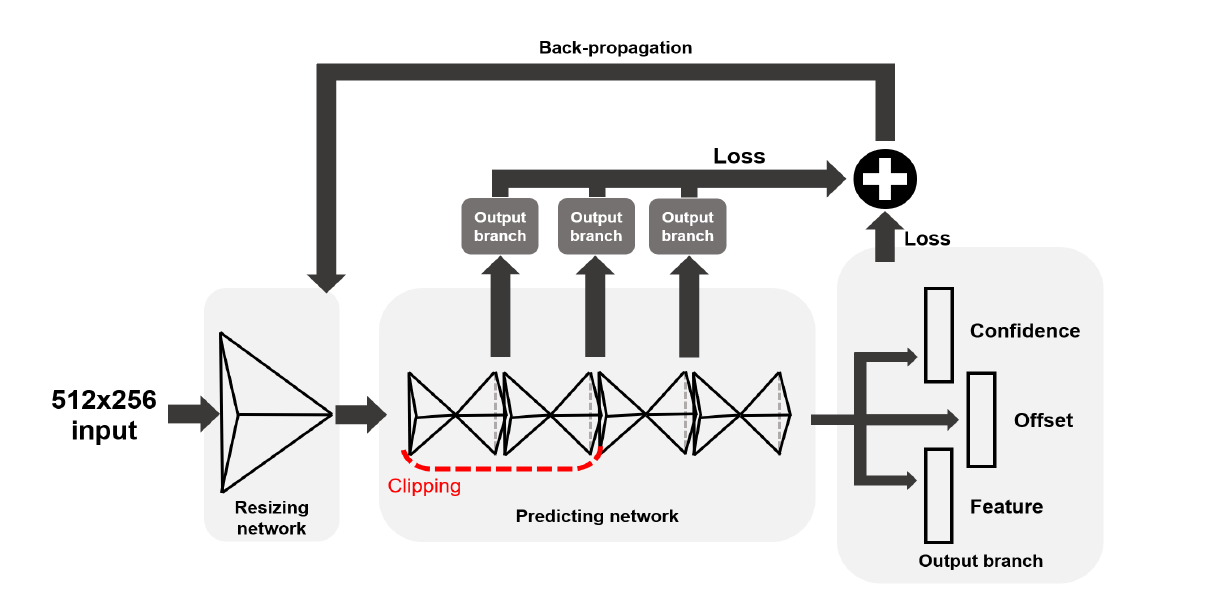}
  \vspace{-1.5em}
  \caption{The overview of PINet \cite{ko2021key}.}\label{fig:PINet structure}
  \vspace{-1.2em}
\endminipage\hfill
\minipage{0.5\textwidth}
  \includegraphics[width=\linewidth]{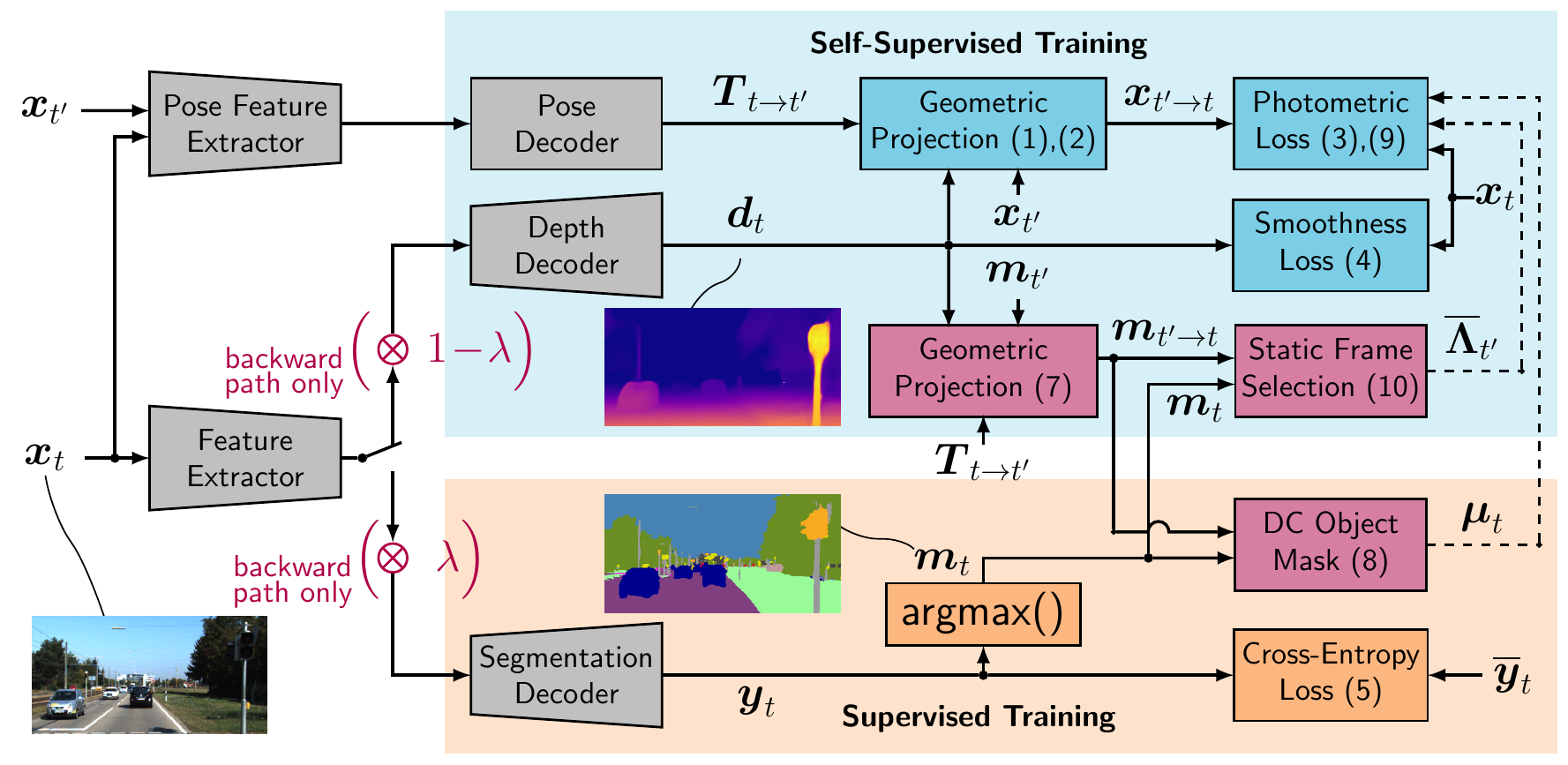}
  \vspace{-1.5em}
  \caption{The overview of SGDepth
  \cite{sgdepth}.}\label{fig:sgdepth_overview}
  \vspace{-1.2em}
\endminipage
\end{figure}

\vspace{-0.75em}
The Point Instance Network (PINet) detects traffic lines regardless of their number \cite{ko2021key}. It generates points on lanes and separates them into distinct instances. As illustrated in Fig \ref{fig:PINet structure}, three output branches are included in this network: a confidence branch, an offset branch, and an embedding branch. Predicting the exact points of traffic lines is what the confidence and offset branches do. The embedding branch creates the embedding features of the predicted points, which are provided in the clustering process to differentiate each instance. First, the input RGB image enters the resizing network, and the sequence of three convolution layers compresses the image into a lower size. After each convolution layer, PReLU \cite{he2015delving} and Batch Normalization \cite{ioffe2015batch} are used. Then, the predicting network receives the resizing network output. It predicts the exact points on the traffic lines as well as embedding features. There are multiple hourglass modules in this network, each with an encoder, decoder, three output branches, and some skip-connections. The predicting network can include any number of hourglass modules, and all of them are trained concurrently by the same loss function. Therefore, in the case of running the trained model on a system with limited computing power, the network can be cut and transferred without extra training. Hourglass blocks have three types of bottlenecks: same, down, and up bottlenecks. The output of the same bottleneck is the same size as the input. In the encoder, the down bottleneck is used for down-sampling, with the first layer of a convolution layer, and in the up-sampling layer, a transposed convolution is used for the up bottleneck. Each output branch has its own channel (confidence: 1, offset: 2, embedding: 4). The associated loss function is applied based on the output branch's goal, and each confidence output is passed on to the following block.
The PINet network loss function equation consists of four different loss functions. Three of them (confidence, offset, and feature loss functions) are parallel and are applied to the network output branch. The other one (distillation loss function) optimizes the knowledge-learning process from the deepest hourglass, which is beneficial if the prediction network is cut to a lighter one. The total loss function equals the weighted sum of the above four loss functions, as shown below:
\begin{equation}
    \begin{aligned}
        L_{total}= & \; \gamma_e L_{exist} + \gamma_n L_{non-exist} + \gamma_o L_{offset} + \gamma_f L_{feature} + \gamma_d L_{distillation} .
    \end{aligned}
\end{equation}
Where constant coefficients are obtained experimentally.

\vspace{-0.75em}
\subsection{YOLO}
\vspace{-0.75em}
 YOLOv5 \cite{glenn_jocher_2022_6222936} network is a single-stage detector which consists of three vital parts. The backbone, the initial element, extracts image features at various scales. The neck, which merges the retrieved features, is the second element. And the network head predicts the box and class of each object in the picture using the features coming from the neck. Eventually, Yolo encodes all the information into a single tensor for each image \cite{yolo}. YOLO models the detection problem as regression by dividing the input image into an $S \times S$ grid. Each grid cell predicts $B$ bounding boxes with $x$, $y$, $w$, and $h$, and an ``objectness'' score $P(Object)$, which indicates whether or not the grid cell contains an object. In addition, a conditional probability $P(Class \mid Object)$ is predicted for the class of the object associated with each grid cell. Therefore, YOLO outputs $B \times 5$ parameters and $C$ class probabilities for each grid cell. These predictions are represented by a tensor with the size $S\times S \times (B \times 5 + C)$. Non-Maximum Suppression (NMS) and thresholding are the last steps in generating final object detection predictions \cite{yolo}. The network parameters are trained by minimizing a three-element loss function called GIoU, Objectness, and Classification. The GIoU element minimizes bounding box prediction error, and the objectness element determines the existence of an object in a grid cell. Finally, the classification error is responsible for object class prediction. 
 YOLOv5 benefits from previous version features, but it is implemented in PyTorch rather than Darknet, making it more flexible with computer vision libraries. 
\vspace{-0.75em}
\subsection{SGDepth}
\vspace{-0.75em}
SGDepth is a novel monocular self-supervised approach to estimate depth information from single images. This process is semantically-guided, meaning it utilizes information obtained from segmentation maps and does not require depth labels. As shown in Fig. \ref{fig:sgdepth_overview}, the approach consists of two main components: the depth part, which is trained in a self-supervised fashion, and the segmentation part, which utilizes a supervised training scheme. For models to estimate depth maps from sequences of images, the world must be static, i.e., in two consecutive frames of a video, all objects must remain in their absolute positions. Moving dynamic-class (DC) entities, including passing vehicles and pedestrians, violate the static world assumptions. Hence, while being necessary for self-supervised depth estimation, correct projections could not be calculated between sequences of frames. In the training phase, SGDepth ignores such objects in their optimization.

\textbf{Self-Supervised Monocular Depth Estimation.} Instead of training the model on depth labels in a supervised manner, the predicted depth maps are then considered as geometric properties of the environment to warp the preceding and succeeding frames  $\boldsymbol{x}_{t'}$, with $t' \in \mathcal{T}' = \left\lbrace {t\!-\!1}, {t\!+\!1} \right\rbrace$ to $\boldsymbol{x}_{t}$ at time t. Afterward, a photometric loss $J_t^{\mathrm{ph}}$ is computed between $\boldsymbol{x}_{t}$ and $\boldsymbol{x}_{t'\rightarrow t}$; this is to ensure that the transformed images $\boldsymbol{x}_{t'\rightarrow t}$, are as close as possible to $\boldsymbol{x}_{t}$. On top of that, a smoothness loss is responsible for making sure that nearby pixels have similar depth values \cite{Godard2017,Godard2019}.

\textbf{Supervised Semantic Segmentation.} A segmentation mask $\boldsymbol{m}_t\in \mathcal{S}^{H\times W}$,  where $\mathcal{S}$ is a set of classes, is obtained by assigning a class to each pixel coordinate. This is done through a supervised training method, and the segmentation head (Fig. \ref{fig:sgdepth_overview}) outputs $\boldsymbol{y}_t\in \mathbb{I}^{H\times W \times S}$ are compared to the ground truth labels $\overline{\boldsymbol{y}}_t$ using a weighted cross-entropy loss \cite{Paszke2016}.

\textbf{Semantic Guidance.} As shown in Fig. \ref{fig:sgdepth_overview}, there are two decoders attached to one encoder (Feature Extractor block). In the backward propagation stage, the gradients that return from the two decoders are scaled to form $\boldsymbol{g}^{\mathrm{total}}$ that is propagated back into the decoder. This is how multi-task training is done across the two domains. To deal with moving DC objects, projected segmentation maps are calculated using nearest-neighbor sampling. In addition, a DC object mask $\boldsymbol{\mu}_t\in \left\lbrace 0, 1\right\rbrace^{H\times W}$ is computed that has zero values for all pixel coordinates that belong to a DC class $\mathcal{S}_{\mathrm{DC}}$ in either one of the three frames. This mask is then applied to compute a semantically-masked photometric loss. If a DC object has moved in two consecutive frames, the warped semantic $m_{t'\rightarrow t,i}$ and the semantic mask $m_{t}$ will be inconsistent. To measure this, intersection over union (IoU) of DC objects in $m_{t'\rightarrow t,i}$ and $m_{t}$ is calculated and denoted as $\Lambda_{t, t'}$. Then a threshold $\theta_{\Lambda}\in\left[ 0,1\right]$ is used to decide whether a frame is static or dynamic, where photometric loss or masked photometric loss is applied, respectively. After this, the total loss will be a combination of the cross-entropy loss and smoothness loss, as well as the photometric losses.
\vspace{-1.5em}
\section{Method}
\vspace{-1.25em}
As previously mentioned, our modular intelligent system includes four parts. In the lane detection section, a series of operations are carried out to display the road lines on the module screen. This section aims to help drivers keep their position in the lane and notice their deviation from the off-road side. In the object detection section, surrounding objects, such as obstacles on the road and traffic signs, are detected so that the system alerts drivers to route appropriately and follow traffic laws. In the segmentation section, the segments of the sidewalks are shown on the screen during the semantic segmentation process, which aids drivers in precepting their environment. Finally, in the distance measurement part, the distance from the nearby cars is estimated using a monocular depth estimation approach to avoid collisions and accidents.

Each of the four module's parts utilizes a base deep learning model, but some do not have a robust and proper result, necessitating some creativity to solve problems. The method section describes how issues are overcome and what novelties are employed to boost the models' performance and create an intelligent modular system. Fig. \ref{fig:overall_diagramm} illustrates the overall system diagram.

\vspace{-1em}
\subsection{Lane Detection}
\label{Lane-Detection}
\vspace{-.75em}
We have used the PINet model for the lane detection part because, as observed from the experiments done and given in Table \ref{table:Lane detection ablation}, it has demonstrated greater accuracy than other rivals while maintaining the required Frame Per Second (FPS). However, the model requires considerable processing power, which results in a noticeable drop in FPS in the performance of our entire module, so PINet is considered a trade-off between accuracy and FPS. 

\begin{wrapfigure}{r}{7cm}
    \vspace{-1em}
	\includegraphics[width=0.9\linewidth]{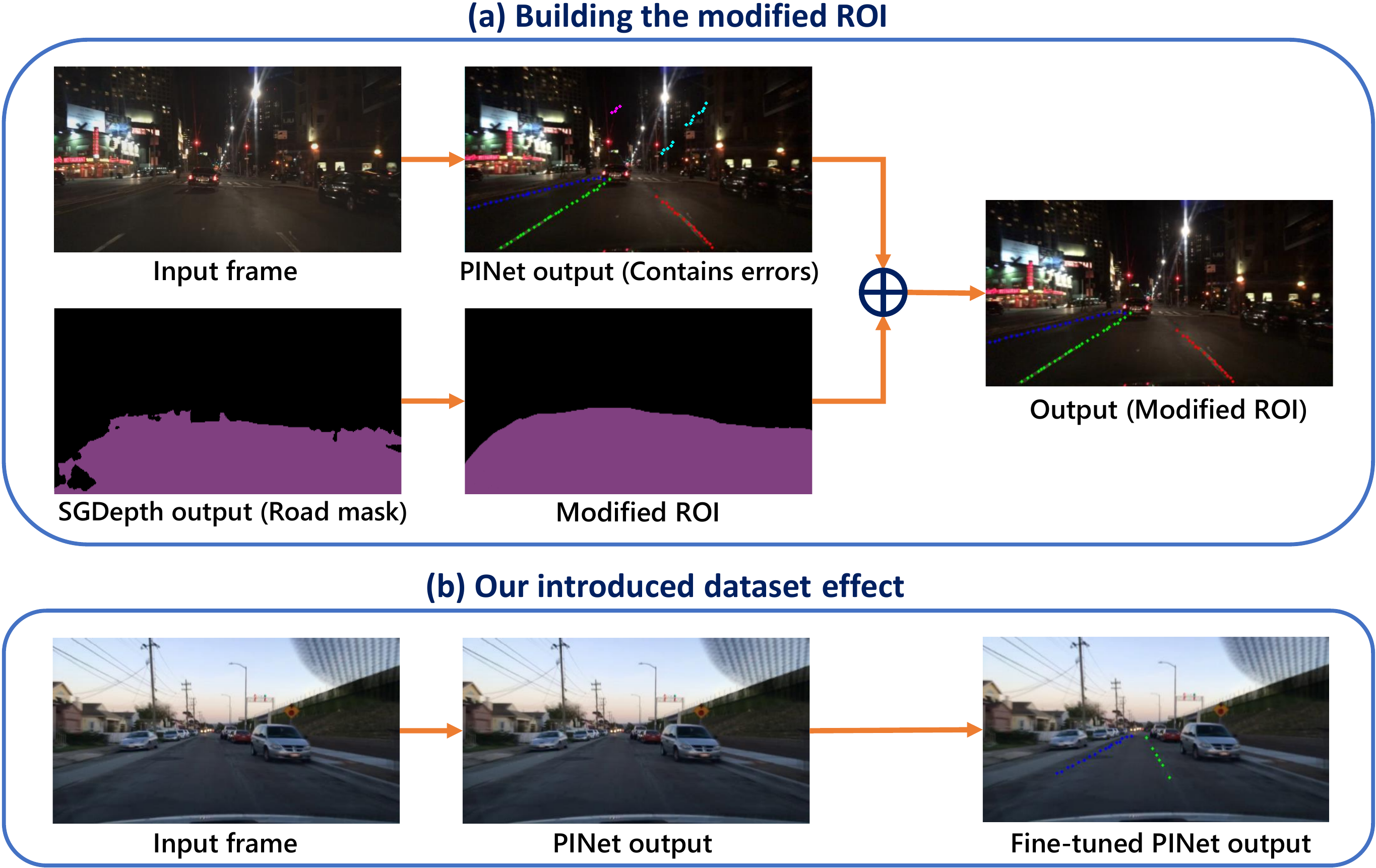}
	\caption{Summary of the implemented method in the lane detection section.}
	\label{fig:lane detection method}
	\vspace{-1.2em}
\end{wrapfigure}

Although PINet shows accurate results for determining road lines, some environmental conditions, like poor illumination, cause severe errors in PINet prediction. In these cases, traffic lines are detected outside the road space. As a result, a novel dynamic region of interest (ROI) is designed and applied to the image to avoid off-road lines being identified. Our experiments show that all issues are addressed after adding this modified ROI to the PINet output, which can be seen in Fig. \ref{fig:lane detection method}.

Building the modified ROI begins with deriving the road mask from the segmentation map output of the SGDepth. The road mask may have some empty areas caused by removing vehicles and objects from the road. Next, the convex hull approach is used on the road segment to obtain a convex space covering all the road mask edges \cite{barber1996quickhull}. After this, a modified and seamless ROI is obtained from the road by filling the convex space gaps. The utilized convex hull formula is as follows:
\begin{equation}
\begin{aligned}
    \text { ROI }=\operatorname{Co}(X)=\left\{\sum_{i=1}^{q} \alpha_{i} x_{i} \mid x_{i} \in X, \alpha_{i} \geqslant 0, \sum_{i=1}^{q} \alpha_{i}=1\right\} 
\end{aligned}
\end{equation}
Where $X=\left\{x_{1}, x_{2}, \ldots, x_{q}\right\}$ denotes a set of points whose covered area is extracted by the convex hull.
Morphological erosion and dilation are also applied for noise canceling and creating soft margins for the mask \cite{haralick1987image}. The erosion of the binary mask by the structuring element B is defined by:
\begin{equation}
    \begin{aligned}
    \text { ROI } \ominus B=\left\{z \in E \mid B_{z} \subseteq \text { ROI }\right\}
    \end{aligned}
\end{equation}
Where $E$ is a Euclidean space, the structuring element $B$ is a circular disc in the plane, and $B_z$  is the translation of $B$ by the vector $z$. Also, the dilation of the binary mask by the structuring element $B$ is defined by:
\begin{equation}
    \begin{aligned}
    \text { ROI } \oplus B=\left\{Z \in E \mid\left(B^{s}\right)_{z} \cap \text { ROI } \neq \emptyset\right\}
    \end{aligned}
\end{equation}
Where $E$ and $B$ are the same as above, and $B^s$ denotes the symmetric of $B$.
Finally, by applying the modified ROI to the input image, the surroundings are removed, and only the in-road detected lines remain, which specify lanes.

Another challenge at this stage is facing roads and ROIs that do not contain traffic lines. To solve this, we introduce a new dataset containing local images collected by us in which roads do not include traffic lines. In the next step, traffic lines are labeled manually, and the network is fine-tuned on a combination of our introduced dataset with pre-existing datasets. As a result, due to the key point extraction nature of PINet, key points, even in the absence of white lines and in noisy conditions, are extracted well, and the fine-tuned PINet is able to divide the road into hypothetical lanes. The dataset introduced in this section is available on the project's GitHub.

\vspace{-1em}
\subsection{Object Detection}
\vspace{-0.5em}
For the object detection part, the YOLO network has been used, which is a state-of-the-art, real-time object detection system. We have chosen the YOLOv5 version because, as shown in Table \ref{table:Object-detection-ablation}, it provides significantly higher accuracy and FPS compared to its counterparts. Utilizing the pre-trained YOLO object detection model, vehicles and pedestrians are detected, and bounding boxes are drawn around them on the monitor to alert the driver. However, identifying traffic signs is challenging since solely detecting them is not sufficient; it is necessary to recognize their meaning so that the driver can follow them. As a result, the YOLO network has been fine-tuned using a combination of multiple traffic signs datasets to both detect them and also identify the type of traffic signs.

The final acquired dataset includes fifteen traffic signs, most of which are similar in shape (e.g., circular with a red margin) but different in meaning. As a result of fine-tuning the model on this dataset, the traffic signs are accurately identified. After a sign is detected, the name of the sign is displayed on the driver's monitor, depending on the type of detected sign. 

Also, for identifying the traffic lights' colors, after using the YOLO and detecting the traffic lights, we use a lightweight CNN-based classifier based on RegNetY002 \cite{radosavovic2020designing} to classify four classes of red, yellow, green, and off states. When a traffic light is detected, the identified color of the light determines the displayed message. If the light is green, the word ``pass'' displays, and if the light is yellow or red, the words ``warning'' or ``stop'' display, respectively.

The results of the object detection section are improved compared to the network's initial version after gathering the datasets and applying the modifications mentioned above, as shown in Table \ref{table:object+detection}.
\vspace{-1em}
\subsection{Segmentation}
\vspace{-0.5em}
We utilize image segmentation for two primary purposes: a) avoiding curb collisions by detecting sidewalks and b) accurately measuring the distance from the vehicle to other objects. 
In the SGDepth model, which is a multi-task learning network, both the tasks of measuring the distance through the monocular depth estimation approach and separating the sidewalk segment from the road are done at the same time. Typically, environmental conditions have a significant influence on the accuracy of the model. Still, by employing this multi-tasking method, the extracted features of both tasks are shared in the network, leading to generalization and better results.

SGDepth is trained on the Cityscapes dataset \cite{Cordts2016Cityscapes}, and we use the sidewalk, car, and bus classes to segment the scene with respect to said class labels. In addition, sidewalks will be colorized and displayed on the monitor. Since the segmentation is usually quite noisy, the approach still needs some novelties and modifications to perform appropriately. We deploy contour analysis and convex hulls to alleviate the sharp edges segmentation map. First, we use morphological functions to dilate the segmented area for each class and fill the holes inside the map that are the outcome of environmental noise. After that, the contours belonging to each semantic class are detected and sorted based on their encapsulating area, and contours whose area is smaller than a constant threshold are discarded. The remaining contours are then converted to convex hulls to compensate for the concavity that might occur in some instances. This way, the imperfections present in the initial segmentation map will be resolved considerably, and a much smoother segmentation map will be obtained. 

\begin{figure*}[t]
\vspace{-1em}
	\centering
	\includegraphics[width=1\textwidth]{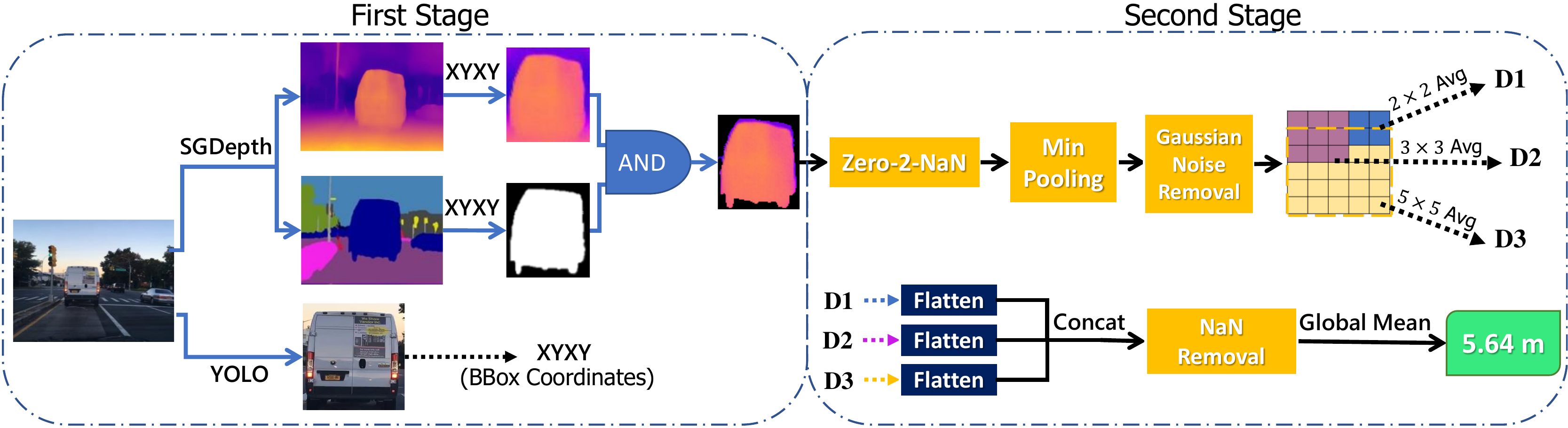}
	\vspace{-1.25em}
	\caption{\textbf{The overview of distance measurement approach.} In the first stage, the object of interest is derived from the input image. Then the distance from the perceived object is determined in the second stage. }
	\label{fig:D_measurement}
	\vspace{-1.5em}
\end{figure*}

\vspace{-1em}
\subsection{Distance Measurement}
\label{section:distance_measurement}
\vspace{-0.75em}
In this section, we propose a novel hierarchical approach that exploits SGDepth and YOLO outputs to attain an accurate distance measurement. Our method consists of two major stages. The first stage is responsible for perceiving the object of interest in the scene. A more in-depth object analysis is achieved in the second stage to obtain the final distance measurement.

Since SGDepth operates on the entire scene, it is vital to identify the surrounding objects first. Therefore, as seen in Fig. \ref{fig:D_measurement}, the object detection output is used to crop the detected object depth map and its segmentation mask based on the bounding box coordinates. We then take advantage of the acquired mask and apply it to the detected depth map to remove the background. 

In order to eliminate the effect of zero-pixel values, we replace them with NaN to not affect the distance measurement calculations. The obtained mask may contain pixels from the background that are assumed to be outliers, and their presence may negatively affect distance measurement. To mitigate their impact, we apply min pooling on the depth map with a kernel size of $3\times3$ in the second stage. Min pooling operation with the kernel size of $K$ is defined as follows
\vspace{-.5em}
\begin{equation}
    \begin{aligned}
        &\acute{I}_{s , p} = min(I_{i+K \times s,j+K \times p})
    \quad\quad\text{Min Pooling}: I^{H\times W} \longrightarrow \acute{I}^{\left\lceil \frac{H}{K} \right\rceil \times \left\lceil \frac{W}{K} \right\rceil}
    \end{aligned}
\end{equation}
for $i=j=0,1,..., K-1$, where $s \in S=\left \{ 0,1,..., \left\lceil \frac{H}{K} \right\rceil -1 \right \}$, and $p \in P=\left \{ 0,1,...,\left\lceil \frac{W}{K} \right\rceil -1  \right \}$ determine the set of new coordinates after applying min pooling. To ensure noisy depth points are not involved in the measurement procedure, we propose a Gaussian Noise Removal module that fits a Gaussian distribution on the depth values. Points whose distances are within an interval of $[\mu -2\sigma, \mu + 2\sigma]$ are known as inliers and remain, while the rest of the points will be excluded and given NaN value to not participate in the distance measurement process. The goal of the Gaussian Noise Removal module is to find inlier points such that
\begin{equation}
    \begin{aligned}
        \text{Inlier Points} = \left \{\acute{I}_{s , p} | \mu -2 \sigma < \acute{I}_{s , p} < \mu + 2\sigma \right \}
        \vspace{-1em}
    \end{aligned}
\end{equation}
where $\sigma$ and $\mu$ denote standard deviation and mean, respectively.
In addition, it is necessary to apply average pooling to the acquired depth map for two main reasons,
(i) to smooth the depth values and ignore the remaining sharp values, and (ii) to achieve an accurate value to report as distance. However, one of the challenging problems of estimating the distance is the object's image size, and applying the prior min pooling would reduce its spatial dimensions by a factor of three. As a result, the kernel size of average pooling should be compatible with the object depth map size to guarantee we have enough depth points for the measurement step. Hence, to satisfy the objects with large and small depth map sizes, we propose a grouped average pooling with different kernel sizes, which are $2\times2$, $3\times3$, and $5\times5$. The output of each average pooling is then flattened and concatenated. The NaN Removal module is subsequently employed to eradicate NaN values. Eventually, the distance to the surrounding objects is obtained by global averaging the resultant depth points. 


\vspace{-1.25em}
\section{Experiments}
\vspace{-0.75em}
For traffic sign detection, we needed both richness and diversity in the signs. To this end, we used a combination of the DFG Traffic Sign Dataset \cite{Vicos_dataset} and Traffic-Sign Detection and Classification in the Wild dataset \cite{Zhe_2016_CVPR}. Furthermore, we used the Common Objects in Context (COCO) dataset \cite{coco_dataset} for the detection of pedestrians and other vehicles. To evaluate the overall performance of our proposed system in all sections, we combined the BDD100K dataset with our collected local dataset. BDD100K \cite{bbd100k} is an extensive collection of 100,000 short videos of driving in various weather and driving conditions. Also, our comprehensive local dataset contains 100 fifteen-second 30 FPS videos of vehicles, roads, traffic signs, pedestrians, and sidewalks. 

\vspace{-1em}
\subsection{Hardware Configuration}
\begin{wrapfigure}{r}{7cm}
	\includegraphics[width=1\linewidth]{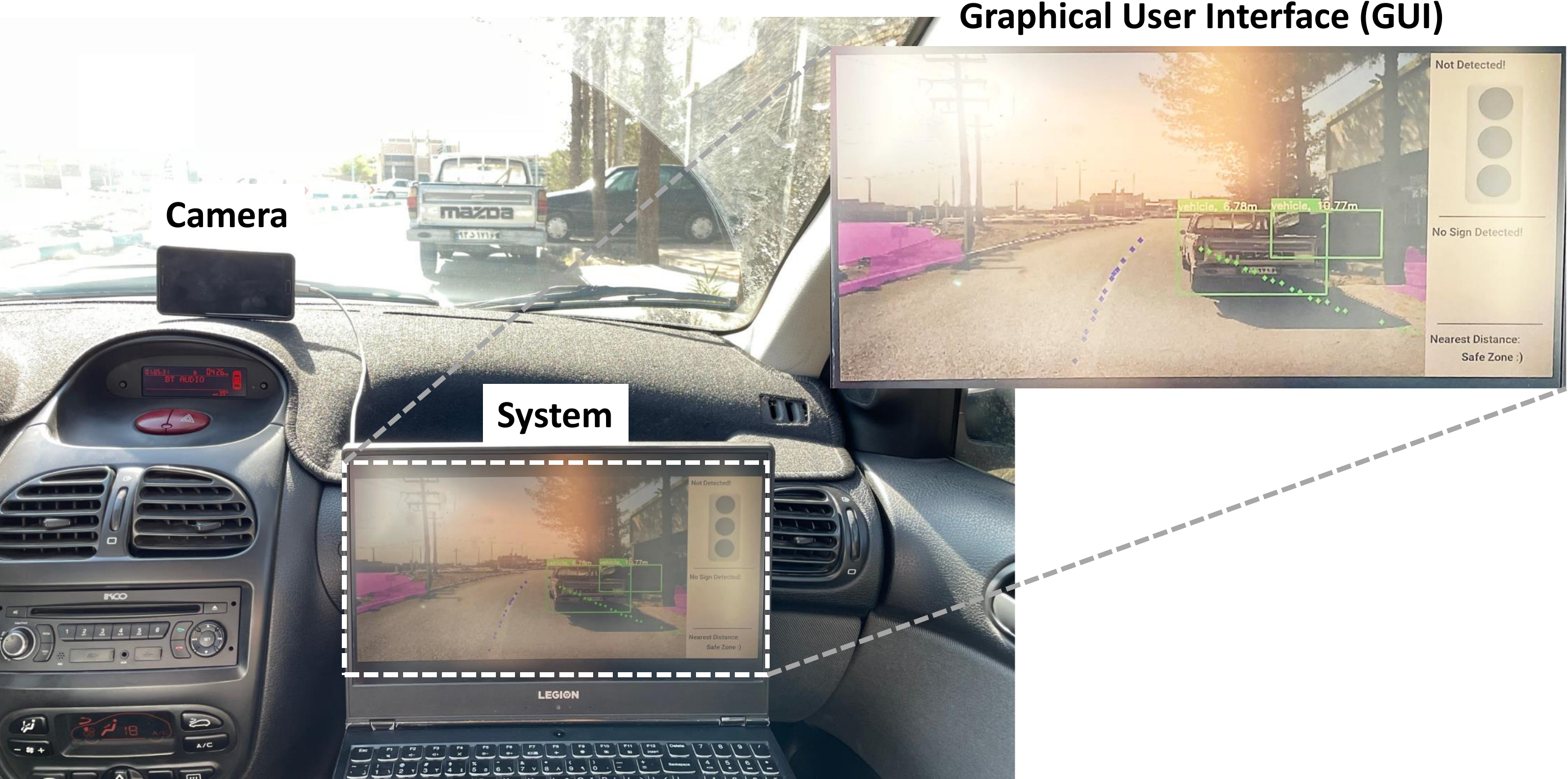}
	\vspace{-1.75em}
	\caption{Implementation of the system in the urban environment. The hardware components include the camera, and the laptop, as a platform for running the package. The output information is displayed to the driver using the GUI.}
	\label{fig:implementation}
	\vspace{-1em}
\end{wrapfigure}

\vspace{-0.5em}
The proposed modular package requires a high-resolution camera, a GPU, and a display. A 1080p camera captures images from the vehicle's perspective, an Nvidia GTX 1660 Ti GPU is in charge of tensor processing in neural networks, and a display indicates the system's results and any necessary alarms to the user. In real-world experiments, some additional hardware components are required for implementation, so a laptop is employed as the hardware framework in the vehicle to avoid this complexity. Fig. \ref{fig:implementation} shows the mentioned system implemented in a standard vehicle.
\vspace{-1.75em}
\subsection{Experimental Results}
\vspace{-0.5em}
Our intelligent system evaluation consists of three phases: assessment of area recognition sections, object detection parts, and distance measurement. Evaluations are performed on 500 frames extracted from the system's output on the BDD100K video dataset and our local dataset. The area recognition section includes detecting sidewalks and traffic lanes. The object detection section comprises identifying pedestrians, parked and moving vehicles, traffic signs, and traffic lights with their colors. Finally, distance labels have been utilized to evaluate the distance measurement section. The assessment methods for each step and the result tables are explained below.
\begin{table}[t]
\tiny\setlength{\tabcolsep}{1pt}
\begin{minipage}{0.28\linewidth}
\centering

\caption{Evaluation of the sidewalk and lane segmentation sections.}
\label{table:segmentation}
\vspace{-1em}
\medskip
\begin{tabular}{clccc}
     &  & \multicolumn{1}{l}{\textbf{IoU}} & \multicolumn{1}{l}{} & \multicolumn{1}{l}{\textbf{Confidence}} \\ \hline
    \multirow{2}{*}{\textbf{Lane}}      & \textit{Local}  & 0.861 & & 91.38\%     \\
                                        & \textit{BDD100K} & 0.842 & & 89.11\% \\ \hline
    \multirow{2}{*}{\textbf{Sidewalk}}  & \textit{Local}  & 0.795 & & 85.34\%       \\
                                        & \textit{BDD100K} & 0.762 & & 80.23\% \\ \hline
\end{tabular}%
\end{minipage}\hfill
\begin{minipage}{0.38\linewidth}
\centering

\caption{Evaluation of the object detection section.}
\label{table:object+detection}
\vspace{-1em}
\medskip
\begin{tabular}{clcccc}
     &  & \multicolumn{1}{l}{\textbf{Precision}} & \multicolumn{1}{l}{\textbf{Recall}} & \multicolumn{1}{l}{\textbf{F1-Score}} & \multicolumn{1}{l}{\textbf{Accuracy}} \\ \hline
    \multirow{2}{*}{\textbf{Vehicles}}  & \textit{Local}  & 96.43\% & 85.71\% & 90.75\% & 83.08\% \\
    & \textit{BDD100K} & 94.92\% & 85.62\% & 90.03\% & 81.87\% \\ \hline
    \multirow{2}{*}{\textbf{Pedestrians}}  & \textit{Local}  & 95.24\% & 90.91\% & 93.02\% & 86.96\% \\
    & \textit{BDD100K} & 92.59\% & 89.28\% & 90.91\% & 83.33\% \\ \hline
    \multirow{2}{*}{\textbf{Traffic Signs}} & \textit{Local}  & 93.33\% & 87.5\% & 90.32\% & 82.35\% \\
    & \textit{BDD100K} & 88.88\% & 94.12\% & 91.43\% & 84.21\% \\ \hline
    \multirow{2}{*}{\textbf{Traffic Lights}} & \textit{Local}  & 94.74\% & 90.00\% & 92.31\% & 85.71\% \\
    & \textit{BDD100K} & 93.55\% & 85.29\% & 89.23\% & 80.56\% \\ \hline
\end{tabular}%
\end{minipage}\hfill
\vspace{-1em}
\begin{minipage}{.28\linewidth}
\centering

\caption{Evaluation of the distance measurement section.}
\label{table:distance}
\vspace{-1em}
\medskip
\begin{tabular}{clcccc}
 &  & \multicolumn{1}{l}{\textbf{mean RA}} &  \multicolumn{1}{l}{ } &  \multicolumn{1}{l}{ } & \multicolumn{1}{l}{\textbf{Accuracy}} \\ \hline
 \multirow{2}{*}{\textbf{Distance}}        & \textit{Local}  & 84.30\% & & & 88.37\% \\ & \textit{BDD100K} & 78.04\% & & & 82.91\% \\ \hline
\end{tabular}
\end{minipage} 
\vspace{-1em}
\end{table}

\noindent\textbf{Lane and sidewalk segmentation.} The segmentation of lane areas and sidewalk is evaluated using the Intersection over Union (IoU) metric, which is the intersection of the predicted and labeled masks per their union. The road lane masks are derived from the regions between the detected lines located in the modified ROI, and the sidewalk masks are the SGDepth modified outputs. These masks are compared to their labels based on the IoU criterion, and a 0.5 threshold is utilized to calculate output accuracy. The predictions with IoU greater than 0.5 are regarded as valid, and with IoU less than 0.5 are considered false. The acquired results are displayed in Table \ref{table:segmentation}.

\noindent\textbf{Object detection.} The performance of our system in object detection is evaluated using precision, recall, F1-score, and accuracy criteria. Notably, detected traffic lights are evaluated as a correct prediction only if their color is classified correctly, as recognizing traffic lights without interpreting their color would not result in getting the right traffic message. The findings are shown in Table \ref{table:object+detection}, which indicates that the system's accuracy is not less than 80\% in any subsection.

\begin{figure}[!t]
\minipage{0.49\textwidth}
	\includegraphics[width=1\linewidth]{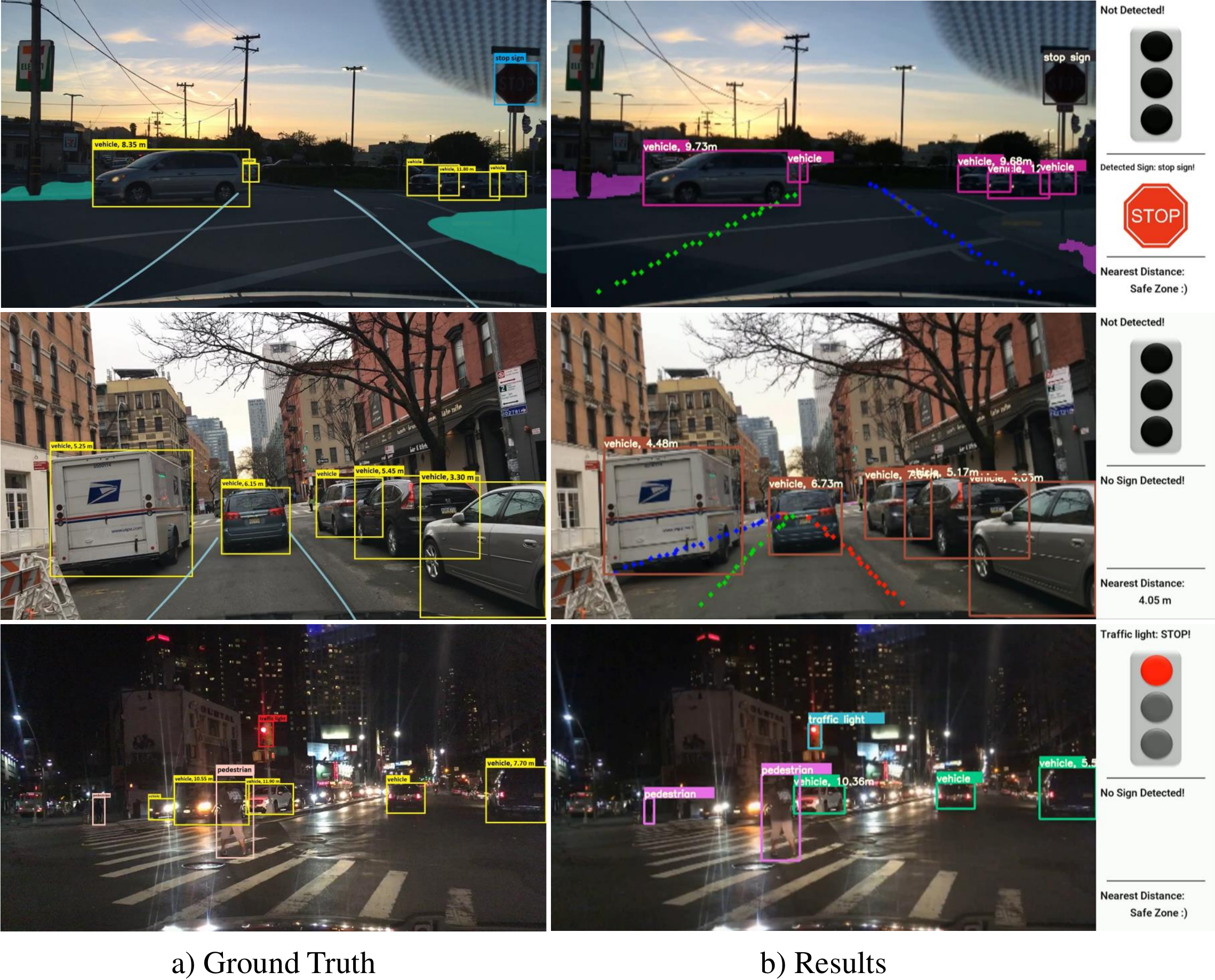}
	\vspace{-1.25em}
	\caption{The proposed system results on the BDD100K dataset.}
	\label{fig:bdd_results}
	\vspace{-1em}
\endminipage\hfill
\minipage{0.49\textwidth}
	\includegraphics[width=1\linewidth]{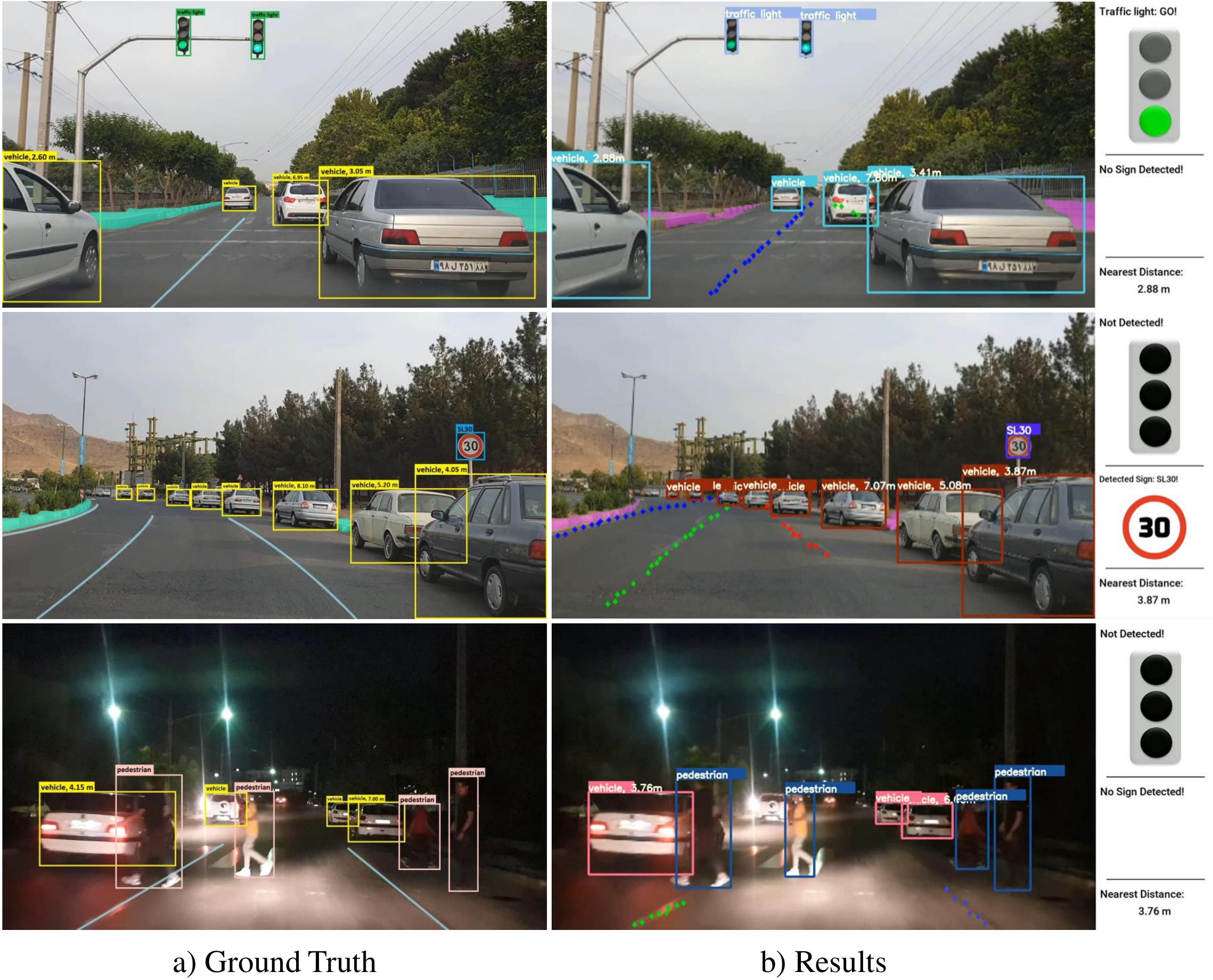}
	\vspace{-1.25em}
	\caption{The proposed system results on the local dataset.}
	\label{fig:local_results}
	\vspace{-1em}
\endminipage\hfill
\end{figure}

\noindent\textbf{Distance measurement.} We utilize relative accuracy (RA) as a metric for the accuracy of the distance measurement section of our proposed system. Relative accuracy is defined using the following formula:
\vspace{-.5em}
\begin{equation}
    \begin{aligned}
        RA = 1 - \frac{\left | Distance_{actual} - Distance_{predicted} \right | }{Distance_{actual}}
        \vspace{-0.5em}
    \end{aligned}
\end{equation}
If $RA$ is greater than 0.8, then the predicted value for distance is considered correct.
Afterward, the accuracy of our system is the percentage of the correct predictions in the entire dataset. The system accuracy and mean relative accuracy are reported for both local and BDD-100k datasets in Table \ref{table:distance}.

Figs.\ref{fig:bdd_results}-\ref{fig:local_results} show the performance results of the proposed perception system on local and BDD100K datasets. A graphical user interface (GUI) is also placed on the right side to represent traffic information to the driver. The GUI information includes traffic light status, traffic signs, and the distance to the nearest car so that the driver can be aware of the situation at a glance.

\begin{table}[t]
\tiny\setlength{\tabcolsep}{2pt}
\begin{minipage}{0.5\linewidth}
\centering
\caption{Real-time Evaluation of the System}
\label{table:real-time}
\begin{tabular}{c|ccccc} 
    \toprule
    Method & \textbf{YOLOv5 (Object)} & \textbf{\textbf{YOLOv5 (Sign)}} & \textbf{PINet} & \textbf{SGDepth} & \textbf{FPS} \\ 
    \hline
    \rowcolor[rgb]{0.824,0.875,0.871} \textbf{Ours} & \checkmark &  \checkmark & \checkmark & \checkmark & 7.56 \\ 
    \hline
    \textbf{A} & \checkmark & \xmark & \xmark & \xmark & 66.06 \\
    \textbf{B} & \xmark & \checkmark & \xmark & \xmark & 69.40 \\
    \textbf{C} & \xmark & \xmark & \checkmark & \xmark & 23.58 \\
    \textbf{D} & \xmark & \xmark & \xmark & \checkmark & 47.48 \\
    \bottomrule
    \end{tabular}
\end{minipage}
\begin{minipage}{0.5\linewidth}
\centering

\caption{Evaluation of different lane detection approaches. NVIDIA Tesla T4 is used for evaluation.}
\label{table:Lane detection ablation}
\begin{tabular}{clcccc}
     &  & \multicolumn{1}{l}{\textbf{F1}} & \multicolumn{1}{l}{\textbf{IoU}} & \multicolumn{1}{l}{\textbf{Confidence}} & \multicolumn{1}{l}{\textbf{FPS}} \\ 
    \hline
    \multirow{2}{*}{\textbf{CLRNet \cite{zheng2022clrnet}}} & \textit{Local} & 48.43\% & 32.05\% & 37.38\% & \multirow{2}{*}{47.55} \\
     & \textit{BDD100K} & 50.90\% & 33.76\% & 39.83\% &  \\ 
    \hline
    \multirow{2}{*}{\textbf{CondLaneNet \cite{liu2021condlanenet}}} & \textit{Local} & 46.52\% & 32.52\% & 38.26\% & \multirow{2}{*}{128.75} \\
     & \textit{BDD100K} & 54.66\% & 37.04\% & 42.61\% &  \\ 
    \hline
    \rowcolor[rgb]{0.824,0.875,0.871} {\cellcolor[rgb]{0.824,0.875,0.871}} & \textit{Local} & 82.57\% & 74.14\% & 82.21\% & {\cellcolor[rgb]{0.824,0.875,0.871}} \\
    \rowcolor[rgb]{0.824,0.875,0.871} \multirow{-2}{*}{{\cellcolor[rgb]{0.824,0.875,0.871}}\textbf{PINet \cite{ko2021key}}} & \textit{BDD100K} & 76.29\% & 75.84\% & 83.55\% & \multirow{-2}{*}{{\cellcolor[rgb]{0.824,0.875,0.871}}24.67} \\
    \hline
\end{tabular}
\end{minipage}
\vspace{-1em}
\end{table}

\vspace{-1em}
\section{Ablation Study}
\vspace{-1em}
\noindent\textbf{Real-time Evaluation of the System.} 
Table \ref{table:real-time} presents the FPS of the whole system and each module individually. Notably, the FPS value for each module is computed from the time spent on preprocessing, model output, and postprocessing. According to Table \ref{table:real-time}, our system achieves a noticeable 7.56 average FPS over BDD100K and local datasets while demonstrating impressive quantitative and qualitative results. A considerable amount of processing time is spent by the PINet model that we have chosen due to its distinctive outputs. Although some other models operate in real-time, they do not have the generalization ability shown by PINet (see Table \ref{table:Lane detection ablation}). In addition, part of the system's processing is utilized to display the UI and model outputs, all of which reduce the processing time of the entire system. Nevertheless, the 7.51 average FPS is a reasonable quantity for running the model in real-time and obtaining satisfactory results. Furthermore, Human Reaction Time (HRT) in driving is the time between when the driver is placed in a critical situation and when he decides to take action \cite{klapp1995motor}. While a constant value for HRT cannot be reported, different methods report values between 1.27s to 1.55 \cite{ranjitkar2003stability, ahmed1999modeling}. We believe that our system is real-time because its overall response time is more than six times less than that of humans. It is possible to connect our module to the vehicle's braking system so that it can slow down in critical situations; this adds the benefit of much faster response times compared to that of humans, rendering our system real-time.

\noindent\textbf{Ablation on Lane Detection} We evaluate different SOTA approaches in lane detection on our proposed local and BDD100k datasets in terms of F1 accuracy, IoU, confidence, and FPS. IoU is first computed between predictions and ground truth, and lanes whose IoU exceeds a threshold (0.5) are considered true positives (TP). It is evident in Table \ref{table:Lane detection ablation} that despite CondLaneNet \cite{liu2021condlanenet} and CLRNet's \cite{zheng2022clrnet} high FPS, their accuracy is significantly inferior to PINet's. Therefore, this gives us the insight that such approaches are not generalizable across different environments. Specifically, they failed when no lines were present in the scene or when other environmental objects obscured the lines. PINet, however, has shown promising results when tested in a new setting, despite some errors that often occur under poor lighting conditions, which we have discussed in \ref{Lane-Detection} on how to improve it. Overall, we have selected PINet as our lane detector since it delivers the requisite FPS for the system's real-time performance while attaining high accuracy in recognizing lanes. 

\noindent\textbf{Ablation on Object Detection.} Table \ref{table:Object-detection-ablation} exhibits the performance of current SOTA approaches for the object detection task on our proposed local dataset and BDD100K. Results indicate that despite the slight difference between approaches, YOLOv5 \cite{glenn_jocher_2022_6222936} outperforms all the latest object detection methods in terms of mAP, mAP, and FPS, making it the best choice for our object detection module. 
\begin{wrapfigure}{r}{0.55\textwidth}
\makeatletter
\def\@captype{table}
\makeatother
\vspace{-1em}
\caption{Evaluation of different object detection approaches. NVIDIA Tesla T4 is used for evaluation.}\label{table:Object-detection-ablation}
\begin{minipage}{0.55\textwidth}
 \centering
\resizebox{0.9\linewidth}{!}{%
\begin{tabular}{clccc}
     &  & \multicolumn{1}{l}{\textbf{mAP}} & \multicolumn{1}{l}{\textbf{mAR}} & \multicolumn{1}{l}{\textbf{FPS}} \\ 
    \hline
    \multirow{2}{*}{\textbf{YOLOX \cite{ge2021yolox}}} & \textit{Local} & 95.90\% & 89.27\% & \multirow{2}{*}{59.6~} \\
     & \textit{BDD100K} & 90.72\% & 96.02\% &  \\ 
    \hline
    \multirow{2}{*}{\textbf{YOLOv3 \cite{redmon2018yolov3}}} & \textit{Local} & 90.76\% & 87.94\% & \multirow{2}{*}{37.3} \\
     & \textit{BDD100K} & 85.84\% & 91.43\% &  \\ 
    \hline
    \multirow{2}{*}{\textbf{YOLOv4 \cite{bochkovskiy2020yolov4}}} & \textit{Local} & 93.33\% & 87.50\% & \multirow{2}{*}{32.1} \\
     & \textit{BDD100K} & 88.88\% & 94.12\% &  \\ 
    \hline
    \rowcolor[rgb]{0.824,0.875,0.871} {\cellcolor[rgb]{0.824,0.875,0.871}} & \textit{Local} & 96.01\% & 90.91\% & {\cellcolor[rgb]{0.824,0.875,0.871}} \\
    \rowcolor[rgb]{0.824,0.875,0.871} \multirow{-2}{*}{{\cellcolor[rgb]{0.824,0.875,0.871}}\textbf{\textbf{YOLOv5 \cite{glenn_jocher_2022_6222936}} }} & \textit{BDD100K} & 92.42\% & 96.28\% & \multirow{-2}{*}{{\cellcolor[rgb]{0.824,0.875,0.871}}90.9} \\
    \hline
\end{tabular}
}
\end{minipage}
\vspace{-1em}
\end{wrapfigure} 
\vspace{-2.1em}
\section{Discussion and Limitations}
\vspace{-1em}
Our experimental results demonstrate the power of our proposed modular system in perceiving the environment and safe driving. The selected networks have been chosen according to the trade-off of accuracy and FPS compared to their counterparts and show high accuracy results. The package FPS can even go higher as the PINet network alone has moderate FPS, and when this network is added to the package, even while the total package stays in real-time, its speed is reduced. Also, we may encounter missing frames on rare occasions, but because it is only one frame in 30 frames, for instance, it is hardly visible. However, by resolving this minor missing frame, the package's accuracy will improve.
In addition, with the development of the package, we can also connect it to the brake and gas pedals. In such a way that after detecting the object at a distance less than a threshold in front of the car, the brake is activated, and the gas pedal is pressed according to the distance from the front cars and the speed limit. It brings us closer to safe driving without the high costs of self-driving vehicles. Overall, the perfect integration of multiple networks and applying appropriate changes to improve their performance, along with the designed GUI, has created a system that gives the necessary warnings to the driver while driving and reduces the risks of driving.
\vspace{-1em}
\section{Conclusions}
\vspace{-1em}
In this paper, we have proposed an intelligent modular vision-based system to assist drivers in safe driving by alerting them in critical moments. There are four main stages in our proposed system: lane detection, object and sign detection, sidewalk segmentation, and distance measurement. PINet is used for lane detection and fine-tuned with a combination of our presented local dataset and BDD100K to alleviate its line-free road estimation issue. To prevent PINet from being impacted by turbulence related to the environment, a novel dynamic ROI is applied to the PINet output. Yolov5 has also been utilized to detect 3 class labels and 15 distinct traffic signs. Moreover, having leveraged SGDepth outputs, monocular depth estimation, and segmentation, we have developed a novel hierarchical method to precisely calculate the distance from neighboring vehicles. In addition, a graphical user interference (GUI) is designed to inform the driver about the traffic light's status, the nearest distance from the adjacent vehicles, and information about traffic signs ahead. Extensive experiments on our local and BDD100K datasets demonstrate that our proposed system performs noticeably well in different environments with different scenarios and conditions and can be an excellent tool to reduce human errors. In the future, we will concentrate on making our system more efficient so that it can operate on embedded boards.

{
\small
\bibliographystyle{plain}
\bibliography{bibfile}
}



\end{document}